\documentclass{article}

\usepackage[preprint]{neurips_2026}

\usepackage[utf8]{inputenc}
\usepackage[T1]{fontenc}
\usepackage{hyperref}
\usepackage{url}
\usepackage{booktabs}
\usepackage{amsfonts}
\usepackage{amsmath}
\usepackage{amssymb}
\usepackage{nicefrac}
\usepackage{microtype}
\usepackage{xcolor}
\usepackage{graphicx}
\usepackage{wrapfig}
\usepackage{capt-of}
\usepackage{multirow}
\usepackage{enumitem}
\usepackage{tikz}
\usetikzlibrary{arrows.meta, positioning, calc}

\title{
\textsc{SAGE}: Scalable Automated Robustness Augmentation for LLM Knowledge Evaluation
}

\author{%
  Xiaoyuan Li$^{1}$\thanks{Equal contribution.}  \quad
  Yuzhe Wang$^{2}$\footnotemark[1]  \quad
  Moxin Li$^{3}$ \quad
  Keqin Bao$^{1}$ \quad
  Rui Men$^{2}$ \quad \\
  \textbf{Yichang Zhang}$^{2}$ \quad
  \textbf{Dayiheng Liu}$^{2}$ \quad
  \textbf{Wenjie Wang}$^{1}$ \quad 
  \textbf{Fuli Feng}$^{1}$ \quad \\
  $^{1}$University of Science and Technology of China \quad
  $^{2}$Alibaba Group \quad \\
  $^{3}$National University of Singapore \\
}

\begin{document}

\maketitle

\begin{abstract}

Large Language Models (LLMs) achieve strong performance on standard knowledge evaluation benchmarks, yet recent work shows that their knowledge capabilities remain brittle under question variants that test the same knowledge in different forms.
Robustness augmentation of existing knowledge evaluation benchmarks is therefore necessary, but current LLM-assisted generate-then-verify pipelines are costly and difficult to scale due to low-yield variant generation and unreliable variant verification.
We propose \textbf{SAGE} (\textbf{S}calable \textbf{A}utomated \textbf{G}eneration of Robustness B\textbf{E}nchmarks), a framework for scalable robustness augmentation of knowledge evaluation benchmarks using fine-tuned smaller models.
SAGE consists of VariantQual, a rubric-based verifier trained on human-labeled seed data, and VariantGen, a variant generator initialized with supervised fine-tuning and further optimized with reinforcement learning using VariantQual as the reward model.
Experiments on HellaSwag show that SAGE constructs a large-scale robustness-augmented benchmark with quality comparable to the human-annotated HellaSwag-Pro at substantially lower cost, while the fine-tuned models further generalize to MMLU without benchmark-specific fine-tuning.

\end{abstract}

\section{Introduction}
\label{sec:intro}


Large Language Models (LLMs) encode substantial knowledge, which enables them to perform well across a wide range of tasks. Evaluating the knowledge capabilities of LLMs is therefore critical for their trustworthy deployment.
Knowledge evaluation benchmarks such as CommonsenseQA~\citep{talmor2019commonsenseqa} and MMLU~\citep{DBLP:conf/iclr/HendrycksBBZMSS21} have become widely used standards for LLM evaluation.
Current state-of-the-art LLMs, including GPT-4o~\citep{openai2023gpt4} and Claude-3.5~\citep{anthropic2024claude}, achieve over 90\% accuracy on these benchmarks, demonstrating strong knowledge capabilities under standard evaluation settings.

Despite near-saturated performance on standard benchmarks, recent work shows that LLMs remain brittle to question variants~\citep{li2025hellaswagpro}, which preserve the tested knowledge but alter the formulation or reasoning process of the original question. 
For example, GPT-4o achieves about 90\% accuracy on the original HellaSwag~\citep{zellers2019hellaswag} but drops to 9\% under negative transformations, \textit{i.e.}, from ``The lady \emph{will}\ldots'' to ``The lady \emph{will not}\ldots'', revealing a serious but often hidden weakness in LLM knowledge capabilities. 
These findings highlight the need for robustness augmentation of existing knowledge evaluation benchmarks with such question variants, enabling robust and comprehensive evaluation of LLM knowledge capabilities.

However, robustness augmentation of large-scale knowledge evaluation benchmarks is both costly and challenging.
Existing efforts such as HellaSwag-Pro~\citep{li2025hellaswagpro} typically adopt an LLM-assisted generate-then-verify pipeline, where strong LLMs generate variants under predefined variant types and reformulation rules, and then judge their validity to filter out invalid candidates.
This pipeline faces two critical bottlenecks.
First, \textbf{variant generation has low yield}: LLMs often produce invalid or low-quality variants, and only 46\% of generated candidates survived quality control in HellaSwag-Pro, resulting in substantial wasted token cost.
Second, \textbf{variant verification is labor-intensive}: prompted LLMs do not reliably determine whether a variant is valid, requiring extensive human annotation.
These bottlenecks limit the scalability and cost-effectiveness of robustness augmentation.

To address these bottlenecks, we replace the costly pipeline of prompting strong models with fine-tuned smaller models specialized for generation and verification.
Our key idea is to first build a reliable variant verifier, and then use it to improve variant generation.
Specifically, we decompose variant quality into three general rubric dimensions: \textit{type compliance, label correctness, and answer uniqueness}, and fine-tune a small verifier on human-labeled seed data to provide reliable quality judgments.
Building on this verifier, we further build a small variant generator by first initializing it through supervised fine-tuning (SFT) on human-annotated seed examples, and then optimizing it with reinforcement learning (RL), using the fine-tuned verifier as the reward model to encourage high-quality generation.
This design improves cost-efficiency through small-model deployment and improves accuracy through verifier-guided quality control and generator optimization.

In this paper, we propose \textbf{SAGE} (\textbf{S}calable \textbf{A}utomated \textbf{G}eneration of Robustness B\textbf{E}nchmarks), a framework for scalable automated robustness augmentation of knowledge evaluation benchmarks.
SAGE consists of two components: \textbf{VariantGen} for generating question variants and \textbf{VariantQual} for evaluating their quality.
Applying SAGE to HellaSwag~\citep{zellers2019hellaswag}, we construct a robustness-augmented benchmark with quality comparable to HellaSwag-Pro, while requiring substantially lower cost. 
Furthermore, these components generalize to MMLU~\citep{DBLP:conf/iclr/HendrycksBBZMSS21} without benchmark-specific fine-tuning.
Our contributions are three-fold:

\begin{figure}[t]
  \centering
  \includegraphics[width=1.0\linewidth]{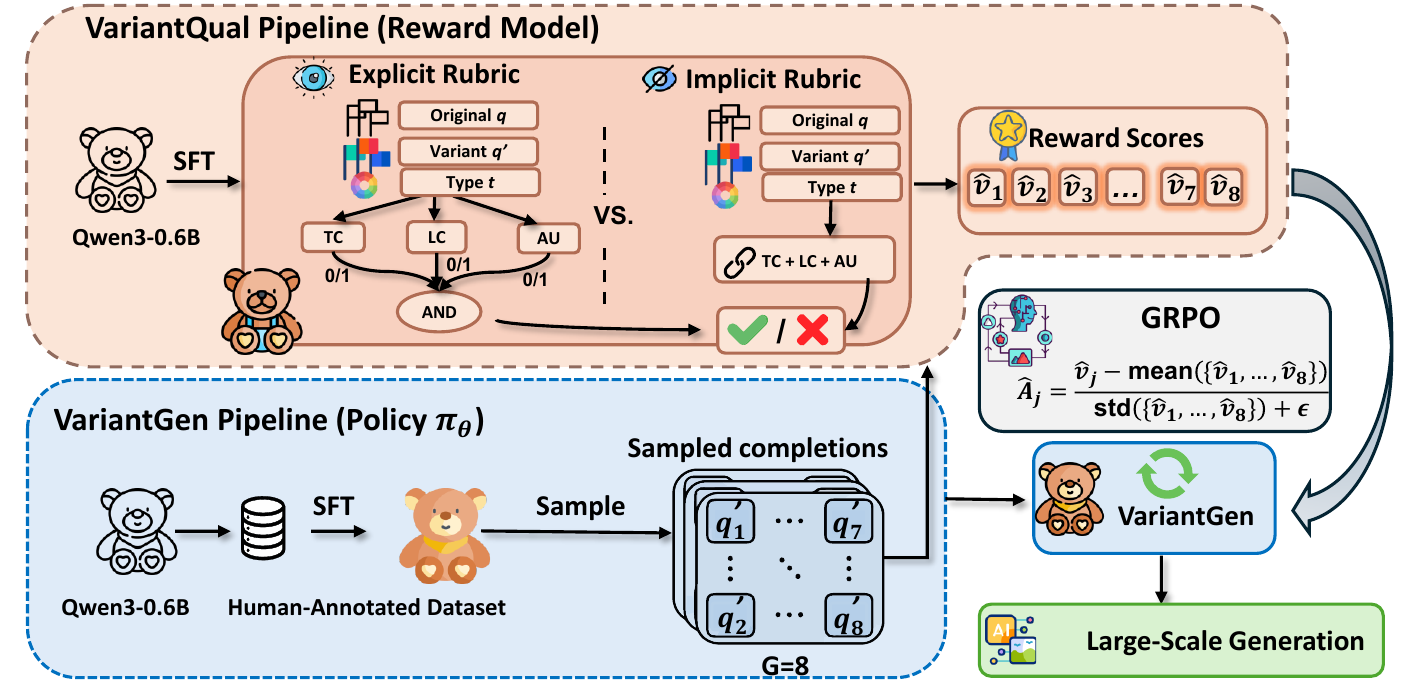}
  \caption{Overview of the SAGE framework. SAGE consists of three stages: SFT of VariantGen and VariantQual on human-annotated dataset, GRPO-based optimization of VariantGen using VariantQual as the reward model, and large-scale benchmark generation with quality filtering.}
  \label{fig:overview}
\end{figure}

\begin{itemize}[leftmargin=*,itemsep=1pt,topsep=2pt]
  \item We formalize the task of scalable automated robustness augmentation for knowledge evaluation benchmarks, which generates question variants under predefined variant types.

  \item We propose SAGE, a framework for scalable automated robustness benchmark augmentation that replaces costly strong-model prompting with fine-tuned small-model generators and evaluators to improve both cost-efficiency and construction accuracy.

  \item We apply SAGE to HellaSwag to construct a large-scale robustness-augmented benchmark with quality comparable to HellaSwag-Pro at substantially lower cost, and further validate its cross-benchmark generalization on MMLU.

\end{itemize}

\section{Related Work}
\label{sec:related}

\paragraph{Robustness and adversarial evaluation.}
Prior work on LLM robustness spans adversarial attacks on reading comprehension~\citep{jia2017adversarial}, universal adversarial triggers~\citep{wallace2019triggers}, spurious correlation analysis~\citep{branco2021shortcutted, geirhos2020shortcut}, and consistency evaluation~\citep{storks2021coherence, johnson2023consistency}. \citet{balepur2024poe} examined negation-based reasoning, while \citet{wu2024counterfactual} studied counterfactual task performance. HellaSwag-Pro~\citep{li2025hellaswagpro} introduced a systematic framework with seven variant types grounded in Bloom's taxonomy, representing the most comprehensive robustness evaluation effort to date. However, all existing robustness benchmarks are manually constructed, limiting their scale and update frequency. SAGE addresses this bottleneck by automating the entire generation and verification pipeline.

\paragraph{Automated data generation.}
Self-Instruct~\citep{wang2023selfinstruct} and Evol-Instruct~\citep{xu2024evolinstruct} demonstrated that LLMs can generate training data for instruction following. \citet{yuan2023distilling} explored constrained language planning with LLM-generated data. However, these methods target general-purpose training data rather than the specific challenge of generating semantically valid, diverse, and adversarially effective robustness evaluation variants. SAGE differs by targeting robustness benchmark generation specifically and employing a rubric-based quality verifier to ensure quality.

\paragraph{Reinforcement learning for text generation.}
RLHF~\citep{ouyang2022rlhf} and preference-based methods such as DPO~\citep{rafailov2023dpo} align LLMs with human preferences using holistic reward signals. GRPO~\citep{shao2024grpo} has proven particularly effective in verifiable domains like mathematics and code, where correctness can be automatically checked to provide clear reward signals. More recently, \citet{gunjal2026rubrics} extended RL to non-verifiable domains by using structured rubrics as reward functions, showing that decomposing quality into explicit criteria can guide optimization where binary verification is unavailable. Inspired by this insight, SAGE applies rubric-guided RL to robustness benchmark generation: VariantQual's rubric decomposes variant quality into concrete, assessable dimensions and serves as the reward model for GRPO.

\section{Method}
\label{sec:method}

\subsection{Problem Definition}
We define robustness augmentation for LLM knowledge evaluation benchmarks as the task of generating valid question variants under predefined variant types.
Given an original multiple-choice question $q=(c,\mathcal{O},y)$, where $c$ is the context, $\mathcal{O}=\{o_k\}_{k=1}^{K}$ denotes the answer choices, and $y \in \{1,\ldots,K\}$ is the correct answer index, let $\mathcal{T}$ denote the set of predefined variant types.
For each target type $t \in \mathcal{T}$, the generated variant $q_t=(c_t,\mathcal{O}_t,y_t)$ should remain a valid multiple-choice question that tests the same underlying knowledge as $q$, while changing its formulation or reasoning process according to type $t$.
In our implementation, $\mathcal{T}$ contains seven variant types guided by Bloom's cognitive taxonomy~\citep{krathwohl1973taxonomy}, as summarized in Table~\ref{tab:variant_types}.

Existing work shows that prompting strong LLMs for variant construction remains costly and difficult to scale due to low-yield generation and unreliable verification.
Motivated by these limitations, SAGE replaces repeated strong-model prompting with two fine-tuned small-model components: VariantQual, a rubric-based verifier for evaluating variant quality, and VariantGen, a generator conditioned on the original question and the target variant type.
We first describe these two components and then present the complete benchmark construction pipeline. Figure~\ref{fig:overview} presents an overview of SAGE.

\begin{table}[t]
  \caption{Seven variant types adopted in SAGE, mapped to Bloom's cognitive taxonomy~\citep{krathwohl1973taxonomy}. Each type transforms the original question into a different reasoning form while preserving the underlying knowledge. Refer to Figure~\ref{fig:case_study_legal} for examples. }
  \label{tab:variant_types}
  \centering
  \small
  \begin{tabular}{@{}lll@{}}
    \toprule
    \textbf{Variant Type} & \textbf{Cognitive Level} & \textbf{Description} \\
    \midrule
    Problem Restatement   & Understand & Rephrase context and correct choice; semantics unchanged \\
    Causal Inference       & Understand & Identify the most plausible cause of an event \\
    Reverse Conversion     & Apply      & Infer the context from the outcome \\
    Scenario Refinement    & Apply      & Modify context so a different choice becomes correct \\
    Negative Transformation & Apply     & Introduce negation; least plausible choice becomes correct \\
    Sentence Ordering      & Analyze    & Determine the correct temporal order of shuffled sentences \\
    Critical Testing       & Evaluate   & Remove key details; ``None of the above'' becomes correct \\
    \bottomrule
  \end{tabular}
\end{table}

\subsection{VariantQual}
\label{sec:method:qual}

We identify two key requirements for fine-tuning a reliable verifier: well-defined evaluation criteria and accurate verification based on these criteria.
We therefore derive a three-dimensional rubric for variant verification and fine-tune VariantQual to learn and apply this rubric. 

\paragraph{Rubric Design.}
\label{sec:method:rubric}

By examining the publicly released annotation data from HellaSwag-Pro, we identify three key requirements for $q_t$ to be valid.
We summarize these requirements into three rubric dimensions.
A variant is considered valid only if it passes all three dimensions.

\begin{itemize}[leftmargin=*,itemsep=1pt,topsep=2pt]
  \item \textbf{Type Compliance (TC)}: Whether the generated variant preserves the original tested knowledge while satisfying the constraints of the target variant type.
  \item \textbf{Label Correctness (LC)}: Whether the annotated label correctly identifies the right answer given the generated context and answer choices.
  \item \textbf{Answer Uniqueness (AU)}: Whether the generated variant has exactly one correct answer, with the other choices being neither duplicate nor irrelevant to the question.
\end{itemize}

We transform the verification annotations from HellaSwag-Pro into seed data for training VariantQual:
\[
\mathcal{D}_{\mathrm{qual}} =
\{(q, q_t, t, l_{\mathrm{TC}}, l_{\mathrm{LC}}, l_{\mathrm{AU}}, v)\},
\]
where $l_{\mathrm{TC}}, l_{\mathrm{LC}}, l_{\mathrm{AU}} \in \{0,1\}$ are dimension-level labels, and $v \in \{0,1\}$ is the final validity label.
Since a valid variant must satisfy all three dimensions, we define
\[
v = l_{\mathrm{TC}} \cdot l_{\mathrm{LC}} \cdot l_{\mathrm{AU}}.
\]
The remaining question is how VariantQual should learn and aggregate these rubric dimensions.
We compare two strategies: \emph{explicit} and \emph{implicit} rubric aggregation.

\paragraph{Explicit Rubric Aggregation (ERA).}
\label{sec:method:explicit}

ERA fine-tunes VariantQual to evaluate each rubric dimension separately.
Given a rubric instruction $r_d$ for dimension $d \in \{\mathrm{TC}, \mathrm{LC}, \mathrm{AU}\}$, VariantQual predicts the corresponding dimension-level label:
\[
p_{\phi}(l_d \mid q, q_t, t, r_d),
\]
where $\phi$ denotes the parameters of VariantQual.
The SFT objective for ERA is to maximize the likelihood of all dimension-level labels:
\begin{align}
\max_{\phi} \ \mathcal{J}_{\mathrm{ERA}}(\phi)
=
\mathbb{E}_{\mathcal{D}_{\mathrm{qual}}}
\left[
\sum_{d \in \{\mathrm{TC}, \mathrm{LC}, \mathrm{AU}\}}
\log p_{\phi}(l_d \mid q, q_t, t, r_d)
\right].
\label{eq:era}
\end{align}
At inference time, VariantQual predicts $\hat{l}_{\mathrm{TC}}$, $\hat{l}_{\mathrm{LC}}$, and $\hat{l}_{\mathrm{AU}}$ separately, and aggregates them as
\[
\hat{v}
=
\hat{l}_{\mathrm{TC}}
\cdot
\hat{l}_{\mathrm{LC}}
\cdot
\hat{l}_{\mathrm{AU}}.
\]
This strategy provides fine-grained diagnostic signals, since each invalid variant can be traced to specific failed dimensions.

\paragraph{Implicit Rubric Aggregation (IRA).}
\label{sec:method:implicit}

IRA fine-tunes VariantQual to directly predict the final validity label by evaluating all rubric dimensions jointly in a single verification pass:
\[
p_{\phi}(v \mid q, q_t, t, \mathcal{R}),
\]
where $\mathcal{R} = [r_{\mathrm{TC}}, r_{\mathrm{LC}}, r_{\mathrm{AU}}]$ denotes the complete rubric instruction.
The SFT objective for IRA is
\begin{align}
\max_{\phi} \ \mathcal{J}_{\mathrm{IRA}}(\phi)
=
\mathbb{E}_{\mathcal{D}_{\mathrm{qual}}}
\left[
\log p_{\phi}(v \mid q, q_t, t, \mathcal{R})
\right].
\label{eq:ira}
\end{align}
This strategy reduces inference cost and allows the model to make a joint decision over all validity requirements.
We fine-tune VariantQual under both ERA and IRA, empirically compare them in Section~\ref{sec:exp_variantqual}, and adopt IRA as the default setting in the SAGE pipeline.

\subsection{VariantGen}
\label{sec:method:gen}

After obtaining VariantQual, we now train VariantGen to generate variants that satisfy the rubrics with a two-stage strategy.
First, we apply SFT on human-annotated seed examples from HellaSwag-Pro to initialize VariantGen for $q_t$ generation conditioned on $q$ and $t$. 
Second, we further optimize VariantGen with GRPO~\citep{shao2024grpo}, using VariantQual to provide verification rewards.

\paragraph{Stage 1: SFT.}
\label{sec:method:sft}

We construct the SFT data as
\[
\mathcal{D}_{\mathrm{gen}}=\{(q, t, q_t^{*})\},
\]
where $q_t^{*}$ is a human-accepted variant of the original question $q$ under variant type $t$.
Let $\pi_{\theta}$ denote VariantGen.
The SFT objective is to maximize the likelihood of the accepted variant:
\begin{align}
\max_{\theta} \ \mathcal{J}_{\mathrm{SFT}}(\theta)
=
\mathbb{E}_{(q,t,q_t^{*}) \sim \mathcal{D}_{\mathrm{gen}}}
\left[
\log \pi_{\theta}(q_t^{*} \mid q, t)
\right].
\label{eq:sft}
\end{align}

\paragraph{Stage 2: GRPO Optimization.}
\label{sec:method:grpo}

SFT teaches VariantGen to imitate accepted variants, but does not directly optimize the verifier-defined quality objective.
We therefore apply GRPO~\citep{shao2024grpo} to further optimize VariantGen using VariantQual's validity prediction as the reward signal.
For each input $x=(q,t)$, VariantGen samples a group of $G$ candidate variants from the old policy $\pi_{\theta_{\mathrm{old}}}$:
\[
q'_1, \ldots, q'_G \sim \pi_{\theta_{\mathrm{old}}}(\cdot \mid x).
\]
VariantQual evaluates each candidate by taking the original question, the generated variant, the target variant type, and the complete rubric instruction as input:
\begin{align}
\hat{v}_j = p_{\phi}(q, q'_j, t, \mathcal{R}),
\quad \hat{v}_j \in \{0,1\}.
\label{eq:reward}
\end{align}
We use $\hat{v}_j$ as the reward signal for updating VariantGen. GRPO computes the group-relative advantage by normalizing the rewards within the sampled group:
\begin{align}
\hat{A}_j
=
\frac{
\hat{v}_j - \mathrm{mean}(\{\hat{v}_k\}_{k=1}^{G})
}{
\mathrm{std}(\{\hat{v}_k\}_{k=1}^{G}) + \epsilon
}.
\label{eq:advantage}
\end{align}

VariantGen is then optimized by maximizing the clipped GRPO objective:
\begin{align}
\max_{\theta} \ \mathcal{J}_{\mathrm{GRPO}}(\theta)
=
\mathbb{E}_{x,\{q'_j\}_{j=1}^{G}}
\Bigg[
&\frac{1}{G}
\sum_{j=1}^{G}
\min
\left(
\rho_j(\theta)\hat{A}_j,
\mathrm{clip}(\rho_j(\theta), 1-\epsilon_{\mathrm{clip}}, 1+\epsilon_{\mathrm{clip}})\hat{A}_j
\right)
\nonumber \\
&-
\beta
D_{\mathrm{KL}}
\left(
\pi_{\theta}(\cdot \mid x)
\middle\|
\pi_{\mathrm{ref}}(\cdot \mid x)
\right)
\Bigg],
\label{eq:grpo}
\end{align}
where
\[
\rho_j(\theta)
=
\frac{
\pi_{\theta}(q'_j \mid x)
}{
\pi_{\theta_{\mathrm{old}}}(q'_j \mid x)
}.
\]
Here, $\pi_{\theta_{\mathrm{old}}}$ is the old policy used to sample candidate variants, $\pi_{\mathrm{ref}}$ is the SFT model used as the reference policy, $\epsilon$ is a small constant for numerical stability, $\epsilon_{\mathrm{clip}}$ is the clipping threshold, and $\beta$ controls the KL penalty strength.

\subsubsection{Benchmark Construction}
Finally, we construct the full benchmark through generate-then-filter.
For each source question, VariantGen generates candidates for all variant types, and VariantQual retains only those passing quality verification.
We repeat this process until for the entire benchmark, each variant type reaches $N$ accepted examples.
The same pipeline is applied to HellaSwag and, without additional fine-tuning, to MMLU for cross-benchmark evaluation.

\section{Experiments}
\label{sec:experiments}
In this section, we present a comprehensive evaluation of SAGE. Our experiments are designed to answer four key research questions:
\textbf{- RQ1}: What is the effectiveness of the generator and quality checker framework in SAGE? 
\textbf{- RQ2}: How can the detection accuracy for VariantQual be improved?
\textbf{- RQ3}: How effective is the SAGE-generated benchmark in evaluating model robustness?
\textbf{- RQ4}: What is the performance of SAGE across different tasks?

\subsection{Experimental Setup}
\label{sec:setup}




\paragraph{Training configuration.}
VariantGen and VariantQual are both initialized from Qwen3-0.6B. Table~\ref{tab:trainingdata} summarizes the training and test set sizes for each module in SAGE. All SFT training uses the LLaMA-Factory ~\citep{zheng2024llamafactory} framework with LoRA adapters (rank~8, $\alpha$~=~16).
GRPO training uses ms-swift ~\citep{zhao2025swift} with num generations~=~8 and max completion length~=~1024;
rollout sampling uses temperature~=~0.9 and top-$p$~=~0.9.
Full hyperparameter details are provided in Appendix~\ref{app:hyperparams} (Tables~\ref{tab:hyperparams_sft} and~\ref{tab:hyperparams_grpo}).

\paragraph{Evaluation metrics.}
For VariantQual we report classification accuracy (ACC), AUC, Recall, and F1 on the held-out test set.
For VariantGen we report pass rate: the fraction of generated variants accepted by VariantQual. 
For the downstream robustness benchmark we report four metrics following HellaSwag-Pro: original accuracy (OA), average robust accuracy (ARA), robust loss accuracy (RLA), and consistent robust accuracy (CRA).

\subsection{Overall Performance (RQ1)}
\label{sec:exp_variantgen}

\begin{wrapfigure}{l}{0.48\textwidth}
  \centering
  \small
  \vspace{-2em}
  \captionof{table}{Overall performance of SAGE on VariantGen (\%). Best results in \textbf{bold}.}
  \label{tab:ablation_training}
  \begin{tabular}{@{}lcc@{}}
    \toprule
    \textbf{Training Stage} & \textbf{Base} & \textbf{Instruct} \\
    \midrule
    SFT Only    & 70.58 & 90.53 \\
    GRPO Only   & 65.77 & 77.54 \\
    SFT + GRPO  & 69.52 & \textbf{91.80} \\
    \bottomrule
  \end{tabular}
  \vspace{1em}
  \captionof{table}{Per-variant-type quality comparison between prompted Qwen2.5-Max and SAGE.}
  \label{tab:per_variant_quality}
  \resizebox{0.92\linewidth}{!}{%
  \begin{tabular}{@{}lcc@{}}
    \toprule
    \textbf{Variant Type} & \textbf{Qwen2.5-Max} & \textbf{SAGE} \\
    \midrule
    Causal Inference    & 97.71 & \textbf{100.00} \\
    Critical Testing    & 96.85 & \textbf{100.00} \\
    Negative Transformation    & 85.44 & \textbf{98.75} \\
    Problem Restatement & 98.61 & \textbf{98.75} \\
    Reverse Conversion  & 78.24 & \textbf{88.75} \\
    Scenario Refinement & 45.56 & \textbf{56.25} \\
    Sentence Ordering   & 98.00 & \textbf{100.00} \\
    \bottomrule
  \end{tabular}%
  }
\vspace{-2em}
\end{wrapfigure}

\paragraph{Main results.}
To assess the effectiveness of SAGE, we conduct experiments as shown in Table~\ref{tab:ablation_training}. VariantGen-SFT (Instruct) achieves 90.53\%, establishing a strong baseline for subsequent RL optimization. Adding GRPO further improves the pass rate to 91.80\%, yielding a gain of 1.27\% over SFT alone. In addition, Instruct models consistently outperform their Base counterparts, suggesting that instruction-following priors transfer effectively to structured variant generation.

\paragraph{Per-variant comparison.}
In Table~\ref{tab:per_variant_quality}, we compare SAGE with a prompt-based Qwen2.5-Max baseline that following HellaSwag-Pro generation protocol. SAGE exceeds Qwen2.5-Max on all seven variant types. \emph{Scenario Refinement} remains the most challenging type for both systems, but the margin indicates that SAGE is notably stronger at handling complex semantic transformations.

\subsection{VariantQual Training (RQ2)}
\label{sec:exp_variantqual}

\begin{wrapfigure}{l}{0.48\textwidth}
  \centering
  \vspace{-2.5em}
  \captionof{table}{Comparison of VariantQual models with different sizes.}
  \label{tab:variantqual_metrics}
  \resizebox{\linewidth}{!}{%
  \begin{tabular}{@{}lcccc@{}}
    \toprule
    \textbf{Base Model} & \textbf{ACC} & \textbf{AUC} & \textbf{Recall} & \textbf{F1} \\
    \midrule
    Qwen3-0.6B & \textbf{90.21} & \textbf{79.10} & \textbf{94.50} & \textbf{94.30} \\
    Qwen3-1.7B & 89.23 & 77.70 & 93.70 & 93.70 \\
    Qwen3-4B & 89.32 & 76.70 & 94.20 & 93.80 \\
    \bottomrule
  \end{tabular}%
  }

  \vspace{0.3em}

  \captionof{table}{Comparison of implicit and explicit rubric strategies.}
  \label{tab:checker_comparison}
  \resizebox{\linewidth}{!}{%
  \begin{tabular}{@{}lcccc@{}}
    \toprule
    \textbf{Rubric Strategy} & \textbf{ACC} & \textbf{AUC} & \textbf{Recall} & \textbf{F1} \\
    \midrule
    IRA & \textbf{90.21} & \textbf{79.10} & 94.50 & \textbf{94.30} \\
    ERA & 87.90 & 64.60 & \textbf{97.10} & 93.20 \\
    \bottomrule
  \end{tabular}%
  }

  \vspace{0.3em}

  \captionof{table}{VariantQual performance before and after RL training.}
  \label{tab:rl_stability}
  \resizebox{\linewidth}{!}{%
  \begin{tabular}{@{}lcccc@{}}
    \toprule
    \textbf{VariantQual} & \textbf{ACC} & \textbf{AUC} & \textbf{Recall} & \textbf{F1} \\
    \midrule
    SFT & \textbf{90.21} & \textbf{79.10} & 94.50 & 94.30 \\
    SFT + GRPO & 90.12 & 77.70 & \textbf{94.90} & 94.30 \\
    \bottomrule
  \end{tabular}%
  }
  \vspace{-3em}
\end{wrapfigure}

\paragraph{Model-size comparison.}
Table~\ref{tab:variantqual_metrics} reports VariantQual performance after SFT across different model sizes. We observe that Qwen3-0.6B performs on par with, or slightly better than, the larger models. This result suggests that the VariantQual task is sufficiently constrained for a 0.6B model to capture the relevant decision boundary given expert-annotated training data, making additional model scale unnecessary. Since the smallest model already delivers competitive performance, we adopt Qwen3-0.6B as the default VariantQual backbone for the rest of our experiments. A similar trend also appears in VariantGen, where Qwen3-0.6B (90.53\%) remains competitive with Qwen3-1.7B (88.06\%) and Qwen3-4B (90.02\%).

\paragraph{Rubric design comparison.}
We compare the implicit rubric (IRA) against the explicit rubric (ERA) described in Section~\ref{sec:method} in  Table~\ref{tab:checker_comparison}. IRA achieves higher ACC (90.21\% vs.\ 87.90\%) and AUC (79.10\% vs.\ 64.60\%). One possible reason is that IRA better adapts to the data distribution during training, whereas ERA may over-rely on specific dimensions, leading to degraded overall performance after the logical AND aggregation.

\paragraph{RL stability analysis.}
To further improve VariantQual's performance and robustness, we conduct additional GRPO training on top of its SFT baseline. We evaluate VariantQual's performance before and after this RL training. As shown in Table~\ref{tab:rl_stability}, GRPO yields marginal and inconsistent changes relative to the SFT baseline (ACC: 90.21\% $\to$ 90.12\%; Recall: 94.50\% $\to$ 94.90\%), with no consistent improvement across metrics. We hypothesize that GRPO optimization might induce a degree of reward hacking by exploiting VariantQual's vulnerabilities, which prevents consistent performance gains. Therefore, we adopt the SFT-trained VariantQual in the final system.

\subsection{Robustness Benchmark Evaluation and Analysis (RQ3)}
\label{sec:exp_benchmark}
To evaluate the effectiveness of the SAGE-generated benchmark in assessing model robustness, 
we deploy the full SAGE pipeline on HellaSwag to generate a large-scale robustness benchmark with 16800 samples ($N = 2400$) and evaluate 12 LLMs spanning four model families.

\begin{table}[t]
\centering
\caption{Robustness evaluation of HellaSwag on the SAGE-generated benchmark. 
All reported values include 95\% confidence intervals. Best results are \textbf{bold}.
}
\label{tab:main_results}
\setlength{\tabcolsep}{5pt}
\begin{tabular}{lcccc}
\toprule
\textbf{Model} & \textbf{OA} (\%) $\uparrow$ & \textbf{ARA} (\%) $\uparrow$ & \textbf{RLA} (\%) $\downarrow$ & \textbf{CRA} (\%) $\uparrow$ \\
\midrule
\multicolumn{5}{l}{\textit{Gemma-3}}\\
\quad Gemma-3-4B & 60.79{\tiny$\pm$1.95} & 33.39{\tiny$\pm$0.71} & 27.40{\tiny$\pm$2.08} & 23.15{\tiny$\pm$0.64} \\
\quad Gemma-3-12B & 64.50{\tiny$\pm$1.91} & 33.73{\tiny$\pm$0.71} & 30.77{\tiny$\pm$2.04} & 24.33{\tiny$\pm$0.65} \\
\quad Gemma-3-27B & 66.17{\tiny$\pm$1.89} & 34.41{\tiny$\pm$0.72} & 31.76{\tiny$\pm$2.02} & 25.33{\tiny$\pm$0.66} \\
\midrule
\multicolumn{5}{l}{\textit{Llama}}\\
\quad Llama-3.2-1B & 53.67{\tiny$\pm$1.99} & 31.83{\tiny$\pm$0.70} & 21.83{\tiny$\pm$2.12} & 20.29{\tiny$\pm$0.61} \\
\quad Llama-3.2-3B & 59.54{\tiny$\pm$1.96} & 32.84{\tiny$\pm$0.71} & 26.70{\tiny$\pm$2.09} & 22.57{\tiny$\pm$0.63} \\
\quad Llama-3.1-8B & 64.71{\tiny$\pm$1.91} & 34.11{\tiny$\pm$0.72} & 30.60{\tiny$\pm$2.04} & 24.79{\tiny$\pm$0.65} \\
\quad Llama-3.1-70B & 71.54{\tiny$\pm$1.80} & 35.68{\tiny$\pm$0.72} & 35.86{\tiny$\pm$1.95} & 27.93{\tiny$\pm$0.68} \\
\midrule
\multicolumn{5}{l}{\textit{Qwen3.5}}\\
\quad Qwen3.5-2B & 54.12{\tiny$\pm$1.99} & 32.50{\tiny$\pm$0.71} & 21.62{\tiny$\pm$2.12} & 20.57{\tiny$\pm$0.61} \\
\quad Qwen3.5-4B & 60.50{\tiny$\pm$1.95} & 34.73{\tiny$\pm$0.72} & 25.77{\tiny$\pm$2.08} & 23.83{\tiny$\pm$0.64} \\
\quad Qwen3.5-9B & 66.21{\tiny$\pm$1.89} & 36.73{\tiny$\pm$0.73} & 29.48{\tiny$\pm$2.03} & 26.86{\tiny$\pm$0.67} \\
\quad Qwen3.5-35B-A3B & 71.79{\tiny$\pm$1.80} & 37.60{\tiny$\pm$0.73} & 34.20{\tiny$\pm$1.94} & 29.26{\tiny$\pm$0.69} \\
\midrule
\multicolumn{5}{l}{\textit{GPT}}\\
\quad GPT-4o & \textbf{74.17}{\tiny$\pm$1.75} & \textbf{65.68}{\tiny$\pm$0.66} & \textbf{11.44}{\tiny$\pm$2.27} & \textbf{50.67}{\tiny$\pm$0.72} \\
\bottomrule
\end{tabular}
\vspace{-18pt}
\end{table}

\paragraph{Main results.}

Table~\ref{tab:main_results} reports OA, ARA, RLA, and CRA for tested LLMs. We find that:
(1)~\textbf{Universal performance gap}: Consistent with HellaSwag-Pro, all models exhibit a substantial OA-to-ARA drop, confirming that robustness deficiencies are pervasive across families and scales.
(2)~\textbf{Robustness lags capability}: For open-source LLMs, OA spans 53.67\%--71.79\%, yet ARA is compressed into 31.83\%--37.60\%, showing that scaling yields diminishing robustness returns.
(3)~\textbf{RLA grows with model size}: Within each family, scaling consistently increases RLA (Llama 1B$\to$70B: 21.83\%$\to$35.86\%; Qwen3.5 2B$\to$35B: 21.62\%$\to$34.20\%)---indicating that
capability gains do not translate proportionally into robustness gains.
(4)~\textbf{CRA is uniformly low}: Even the strongest open-source model achieves only 29.26\% CRA, highlighting the severity of the robustness gap.

\begin{figure}[t]
  \centering
  \includegraphics[width=0.8\linewidth]{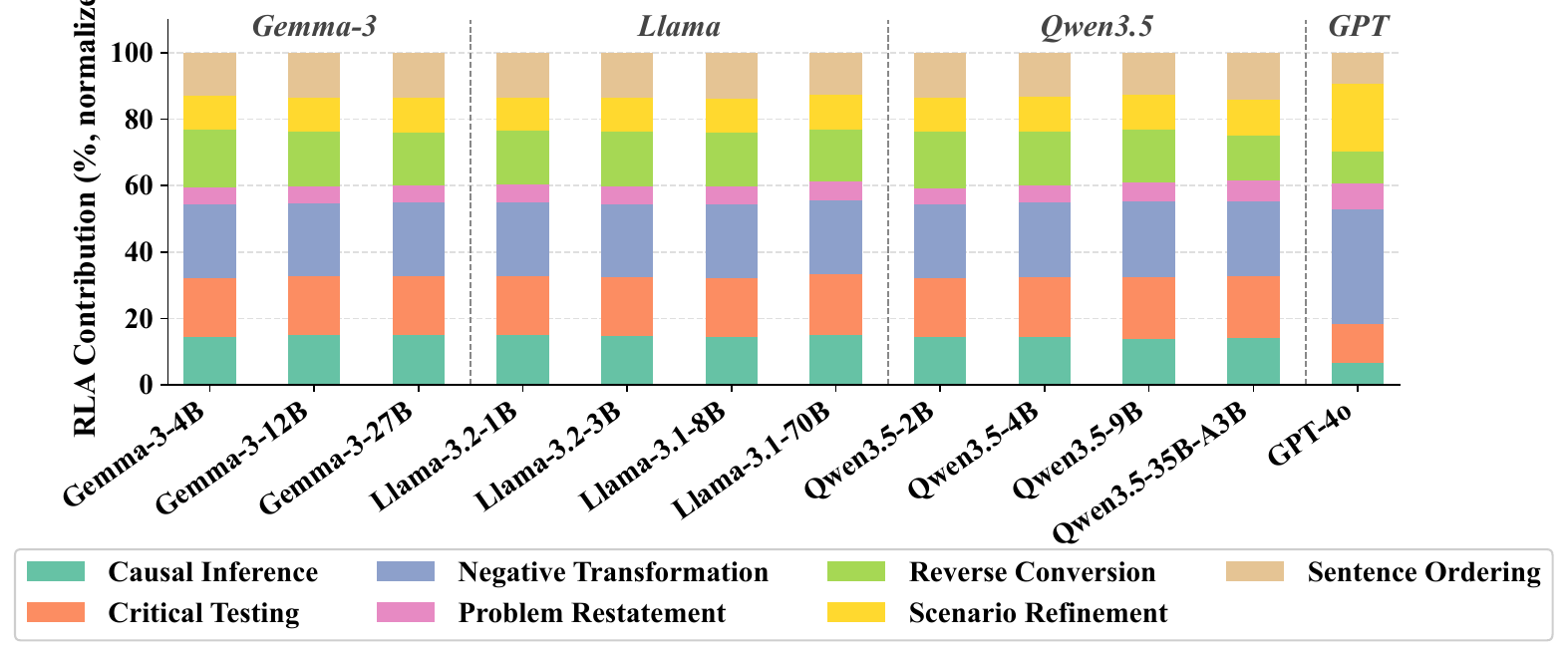}
  \caption{Normalized per-variant-type contribution to RLA across LLMs. }
  \label{fig:rla_contribution_by_type}
\end{figure}

\begin{table}[t]
\centering
\vspace{-10pt}
\caption{Robustness evaluation of MMLU on the SAGE-generated benchmark. 
All reported values include 95\% confidence intervals. Best results are \textbf{bold}.
}
\label{tab:mmlu_main_results}
\setlength{\tabcolsep}{5pt}
\begin{tabular}{lcccc}
\toprule
\textbf{Model} & \textbf{OA} (\%) $\uparrow$  & \textbf{ARA} (\%)$\uparrow$  & \textbf{RLA} (\%)$\downarrow$  & \textbf{CRA} (\%)$\uparrow$  \\
\midrule
\multicolumn{5}{l}{\textit{Gemma-3}}\\
\quad Gemma-3-4B & 65.55{\tiny$\pm$6.04} & 47.78{\tiny$\pm$2.27} & 17.77{\tiny$\pm$6.45} & 34.57{\tiny$\pm$2.22} \\
\quad Gemma-3-12B & 74.79{\tiny$\pm$5.52} & 49.58{\tiny$\pm$2.28} & 25.21{\tiny$\pm$5.97} & 40.16{\tiny$\pm$2.27} \\
\quad Gemma-3-27B & 75.21{\tiny$\pm$5.49} & 49.22{\tiny$\pm$2.29} & 25.99{\tiny$\pm$5.94} & 39.92{\tiny$\pm$2.26} \\
\midrule
\multicolumn{5}{l}{\textit{Llama}}\\
\quad Llama-3.2-1B & 52.10{\tiny$\pm$6.35} & 45.50{\tiny$\pm$2.27} & 6.60{\tiny$\pm$6.74} & 27.91{\tiny$\pm$2.10} \\
\quad Llama-3.2-3B & 62.18{\tiny$\pm$6.16} & 47.12{\tiny$\pm$2.29} & 15.07{\tiny$\pm$6.57} & 32.53{\tiny$\pm$2.20} \\
\quad Llama-3.1-8B & 66.81{\tiny$\pm$5.98} & 47.72{\tiny$\pm$2.28} & 19.09{\tiny$\pm$6.40} & 34.93{\tiny$\pm$2.22} \\
\quad Llama-3.1-70B & 70.17{\tiny$\pm$5.81} & 48.80{\tiny$\pm$2.28} & 21.37{\tiny$\pm$6.24} & 37.21{\tiny$\pm$2.24} \\
\midrule
\multicolumn{5}{l}{\textit{Qwen3.5}}\\
\quad Qwen3.5-2B & 46.22{\tiny$\pm$6.33} & 45.86{\tiny$\pm$2.28} & \textbf{0.36}{\tiny$\pm$6.73} & 24.37{\tiny$\pm$2.02} \\
\quad Qwen3.5-4B & 53.36{\tiny$\pm$6.34} & 46.52{\tiny$\pm$2.29} & 6.84{\tiny$\pm$6.74} & 27.79{\tiny$\pm$2.11} \\
\quad Qwen3.5-9B & 56.72{\tiny$\pm$6.29} & 46.88{\tiny$\pm$2.28} & 9.84{\tiny$\pm$6.70} & 29.95{\tiny$\pm$2.15} \\
\quad Qwen3.5-35B-A3B & 58.40{\tiny$\pm$6.26} & 47.12{\tiny$\pm$2.28} & 11.28{\tiny$\pm$6.66} & 30.49{\tiny$\pm$2.16} \\
\multicolumn{5}{l}{\textit{GPT}}\\
\quad GPT-4o & \textbf{94.12}{\tiny$\pm$2.99} & \textbf{67.33}{\tiny$\pm$1.89} & 28.46{\tiny$\pm$3.03} & \textbf{64.76}{\tiny$\pm$1.94} \\
\bottomrule
\end{tabular}

\vspace{-18pt}

\end{table}

\paragraph{Per-type RLA Contribution.}
To further analyze the impact of different variant types, we calculate the normalized contribution of each variant type to the RLA. 
As shown in Figure~\ref{fig:rla_contribution_by_type}, the distribution is remarkably uniform across models within each family, 
suggesting that variant-type difficulty is a property of the perturbation strategy.
\emph{Negative Transformation} consistently contributes the largest RLA share across all model families. This may be attributed to the presence of statistical shortcuts in pre-training corpora and shortcut learning from corpus-level statistical patterns for this variant type.
In contrast, \emph{Problem Restatement} contributes the smallest RLA share, likely because it involves more surface-level transformations that are easier for models to handle.

Furthermore, to verify that the SAGE-generated benchmark maintains consistency in model ranking with the human-annotated HellaSwag-Pro reference set,
we calculate the Spearman $\rho$ correlation coefficients between these.
This analysis covers OA, ARA, RLA, and CRA across the models shared by both datasets. Among these, ARA achieves $\rho = 0.857$ ($p = 0.0137$) and CRA achieves $\rho = 0.964$
($p = 0.0005$); OA and RLA both achieve $\rho = 1.000$.
These results confirm that SAGE-generated benchmarks faithfully preserve the relative ordering of
model robustness established by human annotation, validating SAGE as a reliable and scalable
alternative to manual benchmark construction. Detailed results are provided in Table~\ref{tab:consistency}.

Meanwhile, we randomly sample 350 SAGE-generated variants and manually verify whether they adhere to the intended constraints, achieving 86.6\% accuracy.
See Appendix~\ref{app:human_eval} for details.

\definecolor{bleudefrance}{RGB}{49,140,231}
\definecolor{cadmiumgreen}{RGB}{0,107,60}
\definecolor{chromeyellow}{RGB}{255,167,0}
\definecolor{darkpastelpurple}{RGB}{150,111,214}
\definecolor{mauvelous}{RGB}{239,152,170}
\definecolor{lightseagreen}{RGB}{32,178,170}
\definecolor{peru}{RGB}{205,133,63}
\begin{figure}[t] 
\centering
\footnotesize
\setlength{\fboxsep}{2pt} 
\fbox{%
\begin{minipage}{0.99\textwidth} 

\textbf{\textsc{Original}} \hfill \textit{MMLU}\\[1pt]
\textbf{Context:} Jacques, Hedwig, and Reyes want to form a business entity through which they can raise capital by selling equity shares to the public. Which of the following business structures should they adopt?\\
\textbf{Choices:} (A)~LLC \quad (B)~PC \quad \underline{(C)~Corporation} \quad (D)~LLP

\vspace{2pt}\hrule\vspace{2pt}

\textbf{\textcolor{bleudefrance}{[1] Problem Restatement}} \hfill \textit{(label: C)}\\
\textbf{Context:} \textcolor{bleudefrance}{Jacques, Hedwig, and Reyes aim to establish a business unit where they can attract investors by offering equity shares. What type of business structure would best suit their needs?}\\
\textbf{Choices:} (A)~LLC \quad (B)~PC \quad \underline{(C)~Corporation} \quad (D)~LLP

\vspace{2pt}\hrule\vspace{2pt}

\textbf{\textcolor{lightseagreen}{[2] Negative Transformation}} \hfill \textit{(label: A)}\\
\textbf{Context:} Jacques, Hedwig, and Reyes want to form a business entity through which they can raise capital by selling equity shares to the public. Which of the following business structures should they \textcolor{lightseagreen}{\textbf{not}} adopt?\\
\textbf{Choices:} \underline{(A)~LLC} \quad (B)~Corporation \quad (C)~Joint-Stock Company \quad (D)~Public Limited Company

\vspace{2pt}\hrule\vspace{2pt}

\textbf{\textcolor{mauvelous}{[3] Scenario Refinement}} \hfill \textit{(label: D)}\\
\textbf{Context:} Jacques, Hedwig, and Reyes want to form a business entity through which they can raise capital by selling equity shares to the public. \textcolor{mauvelous}{However, after consulting with their advisor, they decide to prioritize a partnership-based structure that limits each partner's personal liability for the other partners' actions, rather than pursuing public equity financing.} Which of the following business structures should they adopt?\\
\textbf{Choices:} (A)~LLC \quad (B)~PC \quad (C)~Corporation \quad \underline{(D)~LLP}

\vspace{2pt}\hrule\vspace{2pt}

\textbf{\textcolor{peru}{[4] Critical Testing}} \hfill \textit{(label: E)}\\
\textbf{Context:} Jacques, Hedwig, and Reyes want to form a business entity through which they can raise capital by selling equity shares to the public. \textcolor{peru}{However, regulations in their jurisdiction prohibit any newly formed entity from issuing equity shares to the public within the first five years of registration.} Which of the following business structures should they adopt to meet their immediate goal?\\
\textbf{Choices:} (A)~LLC \quad (B)~PC \quad (C)~Corporation \quad (D)~LLP \quad \underline{\textcolor{peru}{(E)~None of the above four options are suitable.}}

\vspace{2pt}\hrule\vspace{2pt}

\textbf{\textcolor{chromeyellow}{[5] Causal Inference}} \hfill \textit{(label: A)}\\
\textbf{Context:} Jacques, Hedwig, and Reyes want to form a business entity through which they can raise capital by selling equity shares to the public. They should adopt a corporation. \textcolor{chromeyellow}{Which could be the possible reason for this action?}\\
\textbf{Choices:}
\begin{enumerate}[label=(\Alph*), leftmargin=2em, nosep, topsep=2pt]
  \item[\underline{(A)}] \underline{\textcolor{chromeyellow}{A corporation can issue shares of stock to the general public, enabling large-scale capital raising.}}
  \item[(B)] Forming a corporation eliminates all personal tax obligations for the founders.
  \item[(C)] A corporation requires fewer regulatory filings and compliance procedures than other structures.
  \item[(D)] Corporate structures automatically prevent any form of personal liability, including for fraudulent acts.
\end{enumerate}

\vspace{2pt}\hrule\vspace{2pt}

\textbf{\textcolor{cadmiumgreen}{[6] Reverse Conversion}} \hfill \textit{(label: A)}\\
\textbf{Context:} The corporation should be formed by Jacques, Hedwig, and Reyes. \textcolor{cadmiumgreen}{Which could be the possible context for this action?}\\
\textbf{Choices:}
\begin{enumerate}[label=(\Alph*), leftmargin=2em, nosep, topsep=2pt] 
  \item[\underline{(A)}] \underline{\textcolor{cadmiumgreen}{Jacques, Hedwig, and Reyes want to form a business entity through which they}}\\ \underline{\textcolor{cadmiumgreen}{can raise capital by selling equity shares to the public.}}
  \item[(B)] A family business is planning to establish a limited liability company to protect its members' interests.
  \item[(C)] A family business is looking to form a limited liability company to manage its finances more effectively.
  \item[(D)] A family business is planning to form a limited liability company to ensure clear ownership and limited liability.
\end{enumerate}

\vspace{2pt}\hrule\vspace{2pt}

\textbf{\textcolor{darkpastelpurple}{[7] Sentence Ordering}} \hfill \textit{(label: A)}\\
\textbf{Context:} (1)~Jacques, Hedwig, and Reyes want to form a business entity through which they can raise capital by selling equity shares to the public. (2)~They decide to establish a corporation to maximize their investment potential. (3)~Each shareholder elects a board of directors and has voting rights. (4)~The corporation plans to launch its own business operations. \textcolor{darkpastelpurple}{Which is the correct order?}\\
\textbf{Choices:} \underline{\textcolor{darkpastelpurple}{(A)~1--2--3--4}} \quad (B)~2--1--3--4 \quad (C)~1--3--2--4 \quad (D)~4--1--2--3

\end{minipage}%
}
\caption{Case study of SAGE-generated variants from MMLU. The original question and 7 generated variants are shown, with the correct answer choice underlined. }
\vspace{-20pt}
\label{fig:case_study_legal}
\end{figure}

\subsection{Cross-task Evaluation (RQ4)}

\paragraph{MMLU robustness results.}
As shown in Table~\ref{tab:mmlu_main_results}, the robustness degradation pattern observed on HellaSwag transfers consistently to MMLU.
ARA is compressed into 45.50\%--67.33\% despite OA spanning 52.10\%--94.12\%, and scaling increases RLA within each family. 
Compared to HellaSwag, MMLU exhibits smaller RLA values, likely because factual knowledge questions are less susceptible to structural perturbations. Nevertheless, CRA remains moderate, confirming that robustness gaps persist across task domains.

\paragraph{Case study.}
Figure~\ref{fig:case_study_legal} presents a case study of SAGE applied to an MMLU question. SAGE generates seven variants from one question, covering all perturbation types, each adhering to its constraints.
Notably, \emph{Scenario Refinement} shifts the correct answer from Corporation to LLP by introducing a flexibility constraint,
and \emph{Critical Testing} invalidates all original options by adding a liquidity constraint, demonstrating SAGE's ability to generate context-sensitive variants beyond simple paraphrase.

\section{Conclusion}
\label{sec:conclusion}

We proposed SAGE, a scalable framework for automated robustness augmentation of knowledge evaluation benchmarks. SAGE replaces costly strong-model prompting with two fine-tuned small models trained via SFT and verifier-guided GRPO. Applied to HellaSwag, SAGE constructs a 16,800-sample robustness benchmark whose model rankings correlate near-perfectly with human-annotated references, while generalizing to MMLU without additional fine-tuning. We hope SAGE facilitates the shift from static benchmarks toward scalable, continuously updated robustness diagnostics.


\bibliographystyle{plainnat}
\bibliography{references}


\appendix

\section{Hyperparameter Details}
\label{app:hyperparams}

Table~\ref{tab:trainingdata} summarizes the training and test set sizes for each module in SAGE. Tables~\ref{tab:hyperparams_sft} and~\ref{tab:hyperparams_grpo} summarize the hyperparameter configurations used for SFT and GRPO training of VariantGen and VariantQual. 

\begin{table}[htbp]
  \caption{Training and test set sizes for each module in SAGE. 
  Each dataset covers all 7 variant types, which are balanced within each set.
  The GRPO training set is not evaluated on, so no test set is needed.}
  \label{tab:trainingdata}
  \centering
  \small
  \begin{tabular}{@{}lcc@{}}
    \toprule
    \textbf{Module} & \textbf{Train} & \textbf{Test}  \\
    \midrule
    VariantQual                 &10096 & 1124 \\
    VariantGen$_{\text{SFT}}$   &5039  & 561  \\
    VariantGen$_{\text{GRPO}}$  &7000  & -    \\
    \bottomrule
  \end{tabular}
\end{table}

\begin{table}[h]
\centering
\small
\caption{SFT (LoRA) hyperparameters. Both VariantGen and VariantQual share the same configuration.}
\label{tab:hyperparams_sft}
\begin{tabular}{lll}
\toprule
\textbf{Hyperparameter} & \textbf{Value} & \textbf{Description} \\
\midrule
Framework           & LLaMA-Factory & SFT training framework \\
Base model          & Qwen3-0.6B    & Backbone for both modules \\
LoRA rank           & 8             & Intrinsic dimension of LoRA \\
LoRA alpha          & 16            & Scaling factor ($2\times$ rank) \\
LoRA dropout        & 0.0           & Dropout rate on LoRA layers \\
Learning rate       & 1e-4          & AdamW learning rate \\
Temperature         & 0.95          & Sampling temperature during training \\
Max new tokens      & 1024          & Maximum generation length \\
\bottomrule
\end{tabular}
\end{table}

\begin{table}[h]
\centering
\small
\caption{GRPO hyperparameters.}
\label{tab:hyperparams_grpo}
\begin{tabular}{lll}
\toprule
\textbf{Hyperparameter} & \textbf{Value} & \textbf{Description} \\
\midrule
Framework                   & ms-swift  & RLHF training framework \\
Base model                  & Qwen3-0.6B & Backbone initialized from SFT checkpoint \\
Num generations             & 8         & Rollout samples per prompt \\
Max completion length       & 1024      & Maximum generation length per rollout \\
Rollout temperature         & 0.9       & Sampling temperature during rollout \\
Top-$p$                     & 0.9       & Nucleus sampling threshold \\
Learning rate               & 1e-6      & AdamW learning rate \\
$\beta$ (KL penalty)        & 0.04      & KL divergence coefficient \\
Clip range                  & 0.2       & PPO probability ratio clip bound \\
Num train epochs            & 2         & Total training epochs \\
Per-device train batch size & 2         & Batch size per GPU \\
Gradient accumulation steps & 8         & Effective batch size $= 2 \times 8 = 16$ \\
Precision                   & bfloat16  & Mixed-precision training format \\
Gradient checkpointing      & True      & Enabled to reduce GPU memory usage \\
\bottomrule
\end{tabular}
\end{table}

\section{Compute and Cost Details}
\label{app:compute_cost}

The total monetary cost of the reported experiments is approximately 284 USD.
For comparison, the original human annotation process in HellaSwag-Pro cost approximately 7,064 USD, meaning SAGE achieves a $\sim$25$\times$ cost reduction while producing benchmarks of comparable quality.
The main training runs use 15 A40 GPU-hours and 40 A100 GPU-hours, as summarized in Table~\ref{tab:gpu_hours}.

\begin{table}[h]
\centering
\small
\caption{Aggregate cost and GPU-hour usage for the reported experiments.}
\label{tab:gpu_hours}
\begin{tabular}{@{}lc@{}}
\toprule
\textbf{Item} & \textbf{Value} \\
\midrule
Total monetary cost & $\sim$284 USD \\
A40 compute & 15 GPU-hours \\
A100 compute & 40 GPU-hours \\
\bottomrule
\end{tabular}
\end{table}

\section{Ranking Consistency with HellaSwag-Pro}
\label{app:consistency}

We compare model performance on HellaSwag-Pro reference set versus SAGE-generated variants across all four metrics (OA, ARA, RLA, CRA), and report Spearman $\rho$ and Kendall $\tau$ rank correlation coefficients. Full results are shown in Table~\ref{tab:consistency}.

\begin{table}[h]
\centering
\caption{Comparison of model performance on HellaSwag-Pro reference set vs.\ SAGE-generated variants,
with ranking consistency between the two evaluations.
Statistical significance of ranking correlations is assessed via Spearman $\rho$ and Kendall $\tau$ with two-sided $p$-values.}
\label{tab:consistency}
\setlength{\tabcolsep}{4pt}
\begin{tabular}{lcccc|cccc}
\toprule
& \multicolumn{4}{c|}{\textbf{HellaSwag-Pro Reference}} & \multicolumn{4}{c}{\textbf{SAGE Variants}} \\
\textbf{Model} & OA & ARA & RLA & CRA & OA & ARA & RLA & CRA \\
\midrule
\multicolumn{9}{l}{\textit{Gemma-2}}\\
\quad Gemma-2-2B  & 59.62 & 39.13 & 20.50 & 24.88 & 58.29 & 33.43 & 24.86 & 22.43 \\
\quad Gemma-2-9B  & 64.88 & 39.80 & 25.08 & 26.91 & 62.38 & 33.14 & 29.24 & 23.45 \\
\quad Gemma-2-27B & 71.88 & 40.91 & 30.97 & 30.25 & 69.29 & 34.78 & 34.51 & 26.60 \\
\midrule
\multicolumn{9}{l}{\textit{Llama-3}}\\
\quad Llama-3-8B  & 66.25 & 40.21 & 26.04 & 27.34 & 65.38 & 34.26 & 31.12 & 25.13 \\
\quad Llama-3-70B & 72.50 & 41.27 & 31.23 & 30.63 & 71.33 & 35.46 & 35.87 & 27.72 \\
\midrule
\multicolumn{9}{l}{\textit{Mistral}}\\
\quad Mistral-7B-v0.1   & 67.50 & 41.52 & 25.98 & 28.93 & 66.04 & 35.05 & 30.99 & 25.93 \\
\quad Mixtral-8x7B-v0.1 & 69.75 & 41.21 & 28.54 & 29.39 & 69.08 & 35.14 & 33.94 & 26.89 \\
\midrule
\multicolumn{9}{l}{\textit{Ranking Consistency (Reference vs.\ SAGE)}}\\
\quad Spearman $\rho$ & 1.000 & 0.857 & 1.000 & 0.964 & & & & \\
\quad $p$-value        & 0.000 & 0.014 & 0.000 & 0.001 & & & & \\
\quad Kendall $\tau$   & 1.000 & 0.714 & 1.000 & 0.905 & & & & \\
\quad $p$-value        & 0.000 & 0.030 & 0.000 & 0.003 & & & & \\
\bottomrule
\end{tabular}
\end{table}

\section{Human Evaluation of Variant Quality}
\label{app:human_eval}

\begin{figure}[t]
  \centering
  \includegraphics[width=0.8\linewidth]{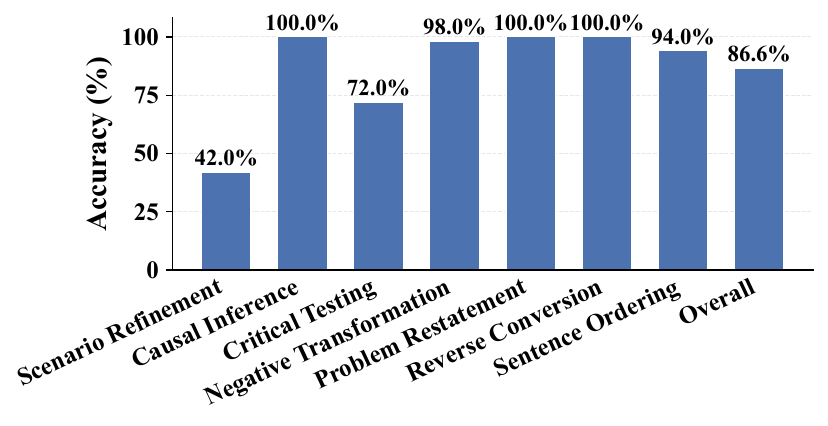}
  \caption{Human verification accuracy of SAGE-generated variants.}
  \label{fig:manual_check_quality_bar}
\end{figure}

To verify that SAGE-generated variants adhere to their intended perturbation constraints, we randomly sample 350 variants (50 per variant type) and conduct manual evaluation. Each variant is assessed on three criteria: (1) perturbation compliance, (2) label correctness, and (3) answer uniqueness.
As shown in Figure~\ref{fig:manual_check_quality_bar}, the vast majority of generated variants successfully satisfy the three criteria above.  This confirms that SAGE effectively produces valid, diverse variants suitable for robust evaluation. 

Although we observe moderately lower strict compliance rates for the \textit{Scenario Refinement} and \textit{Critical Thinking} perturbation types, this is primarily due to the multi-task nature of our training mixture where constraints can occasionally blend. However, even when a variant does not perfectly mirror the strict definition of its assigned perturbation type (e.g., failing to flawlessly "refine" a scenario), it typically still introduces a meaningful and challenging structural modification to the original question. Consequently, these variants remain highly effective for testing reasoning robustness.

\section{Limitations}
\label{app:limitations}

\paragraph{Scope of supported benchmarks and variants.}
SAGE is evaluated on multiple-choice knowledge evaluation benchmarks, with HellaSwag as the primary construction setting and MMLU as a cross-task case study.
The current formulation assumes that each source question has a well-defined correct answer and a fixed set of answer choices.
Consequently, the conclusions may not directly transfer to free-form generation, multi-answer questions, multimodal evaluation, or tasks whose validity depends on long-form explanation rather than selecting one option.
In addition, SAGE generates variants under seven predefined perturbation types.
These types cover a range of reasoning forms, but they do not exhaust all possible robustness failures; extending SAGE to new domains may require designing new perturbation types and collecting a small amount of domain-specific verification data.

\paragraph{Dependence on verifier quality.}
SAGE relies on VariantQual both for filtering generated candidates and for providing rewards during GRPO optimization.
Errors or biases in VariantQual can therefore propagate into the final benchmark, and a generator optimized against this verifier may exploit verifier-specific blind spots.
Although our human evaluation shows high overall quality, it also reveals that some perturbation types, especially Scenario Refinement and Critical Testing, have lower strict compliance rates than simpler transformations.
This indicates that automatic filtering does not fully replace human auditing for the most semantically complex variants.

\paragraph{Interpretation of robustness scores.}
The SAGE-generated benchmark is intended as a diagnostic tool for measuring consistency under controlled question variants, not as a definitive measure of whether a model possesses the underlying knowledge.
Generated variants can change question difficulty in ways that are not perfectly captured by the original-variant pairing, and some variants may introduce artifacts that affect model behavior independently of the intended perturbation.
Therefore, robustness metrics such as ARA, RLA, and CRA should be interpreted together with per-type analyses and manual quality checks.

\paragraph{Data and deployment considerations.}
Because SAGE augments existing benchmarks, it inherits their coverage limitations, annotation artifacts, and potential social or cultural biases.
When applying SAGE to sensitive domains such as medicine, law, finance, or safety evaluation, additional expert review is needed to verify label correctness, fairness, and domain appropriateness.
Moreover, while SAGE reduces reliance on repeated prompting of strong proprietary models, large-scale benchmark construction still requires model fine-tuning, rollout generation, and verification compute.
These costs should be considered when scaling the framework to substantially larger or continuously updated benchmark suites.

\section{Broader Impact}
\label{app:broader_impact}

\paragraph{Positive impact.}
SAGE democratizes the construction of robust evaluation benchmarks by replacing expensive, large-scale human annotation with a lightweight, reproducible pipeline built on small open-source models.
This substantially lowers the financial and logistical barriers to creating high-quality benchmarks, enabling resource-limited research groups and underrepresented communities to participate in LLM evaluation research.
By systematically probing models under controlled perturbations, SAGE encourages the development of more robust and reliable language models, which benefits downstream applications in education, healthcare, legal reasoning, and other high-stakes domains where consistent model behavior is critical.

\paragraph{Potential risks and mitigation.}
Because SAGE can automatically generate large numbers of benchmark variants, there is a risk that the generated variants could be used to overfit models to specific evaluation patterns, thereby undermining the purpose of benchmarking.
We mitigate this by clearly stating that generated variants are intended solely for evaluation, not for training.
Additionally, since SAGE inherits content from existing benchmarks, it may propagate biases present in the source data.
Practitioners applying SAGE to sensitive domains should conduct domain-specific expert review before deployment.
We do not anticipate direct negative societal consequences from the benchmark augmentation methodology itself, as it does not generate harmful content, collect personal data, or enable surveillance capabilities.

\section{Asset Documentation}
\label{app:asset_documentation}

\subsection{Existing Assets}
\label{app:existing_assets}

Table~\ref{tab:existing_assets} summarizes the existing datasets, models, and software frameworks used in this work.
We use these assets for research evaluation and model fine-tuning, cite the corresponding papers or project pages in the main text, and do not redistribute third-party model weights or source datasets beyond their original terms.
For assets whose license or terms are managed by an external provider, users should consult the original resource page before reuse.

\begin{table}[h]
\centering
\small
\caption{Existing assets used in SAGE.}
\label{tab:existing_assets}
\begin{tabular}{@{}p{0.22\linewidth}p{0.29\linewidth}p{0.19\linewidth}p{0.20\linewidth}@{}}
\toprule
\textbf{Asset} & \textbf{Use in this work} & \textbf{License / terms} & \textbf{Source} \\
\midrule
HellaSwag & Source benchmark for large-scale robustness augmentation and evaluation & MIT License & \url{https://github.com/rowanz/hellaswag} \\
HellaSwag-Pro & Human-annotated seed data and reference benchmark for variant generation and verification & CC BY 4.0 & \url{https://arxiv.org/abs/2502.11393} \\
MMLU & Cross-task case study for transfer evaluation & MIT License & \url{https://huggingface.co/datasets/cais/mmlu} \\
Qwen3-0.6B & Backbone for VariantGen and VariantQual & Apache-2.0 & \url{https://huggingface.co/Qwen/Qwen3-0.6B} \\
Qwen2.5-Max & Prompt-based generation baseline via API access & Provider terms & \url{https://www.alibabacloud.com/help/en/model-studio/user-guide/model/} \\
LLaMA-Factory & SFT training framework & Apache-2.0 & \url{https://github.com/hiyouga/LLaMA-Factory} \\
ms-swift & GRPO training framework & Apache-2.0 & \url{https://github.com/modelscope/ms-swift} \\
\bottomrule
\end{tabular}
\end{table}

\subsection{New Assets}
\label{app:new_assets}

This paper introduces two types of new research assets.
First, SAGE generates robustness-augmented multiple-choice benchmark variants from existing source questions.
Each generated example contains the original question identifier, target variant type, generated context, answer choices, and correct label.
The intended use of these generated variants is robustness evaluation of LLM knowledge capabilities, not model training on test answers.
The construction pipeline, variant types, filtering criteria, prompt templates, training data sizes, human evaluation protocol, and limitations are documented in Sections~\ref{sec:method} and~\ref{sec:experiments}, Table~\ref{tab:variant_types}, Appendix~\ref{app:hyperparams}, Appendix~\ref{app:human_eval}, Appendix~\ref{app:limitations}, and Appendix~\ref{app:prompts}.

Second, SAGE trains task-specific VariantGen and VariantQual components initialized from Qwen3-0.6B.
The model documentation includes the backbone model, training frameworks, LoRA configuration, SFT and GRPO hyperparameters, compute usage, and prompt templates in Appendix~\ref{app:hyperparams}, Appendix~\ref{app:compute_cost}, and Appendix~\ref{app:prompts}.
Because the generated benchmark is derived from existing public benchmarks, any release of the derived examples should preserve attribution to the source datasets and follow their licenses or terms.
No additional consent is required for human subjects, since the new assets are generated from public benchmark items and do not contain newly collected personal data.

\section{Prompt Templates}
\label{app:prompts}

\begin{figure}[h]
  \centering
  \includegraphics[width=\linewidth]{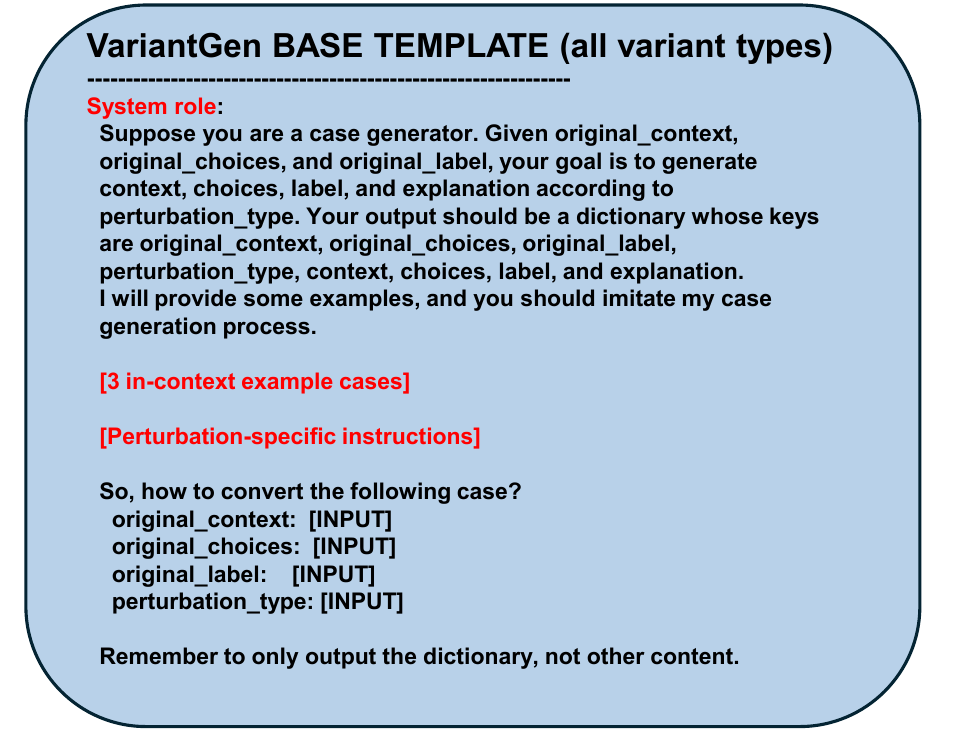}
  \caption{VariantGen base prompt template. The same template is used for all variant types.}
  \label{fig:prompt_generator_base}
\end{figure}

\begin{figure}[h]
  \centering
  \includegraphics[width=\linewidth]{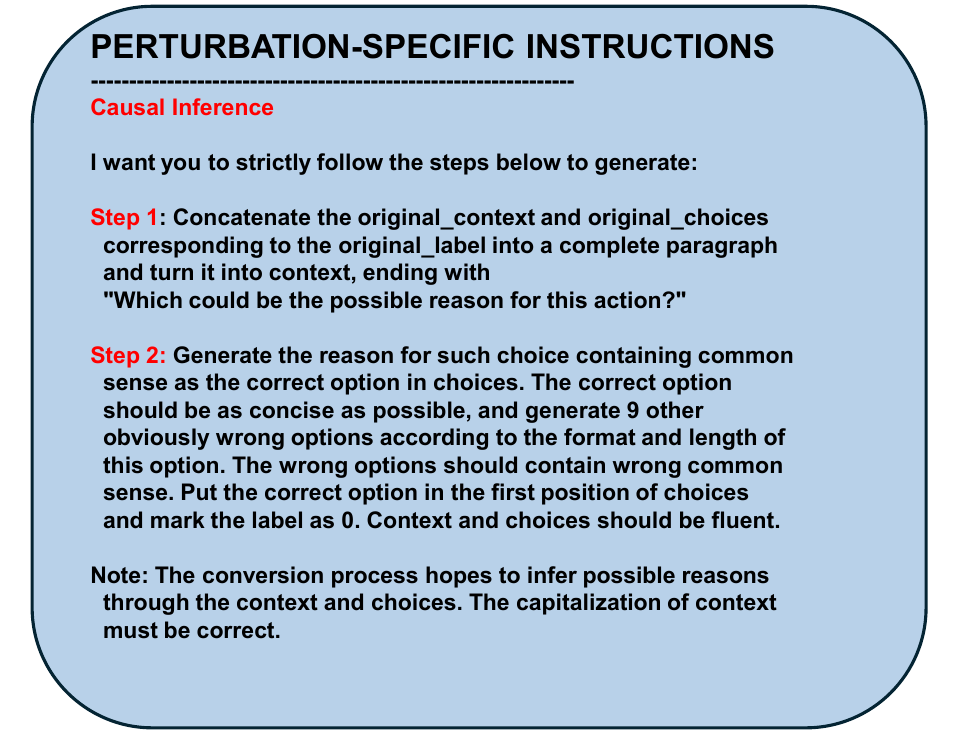}
  \caption{Perturbation instructions for VariantGen. Causal Inference is shown as an example.}
  \label{fig:prompt_generator_causal_inference}
\end{figure}

\begin{figure}[h]
  \centering
  \includegraphics[width=\linewidth]{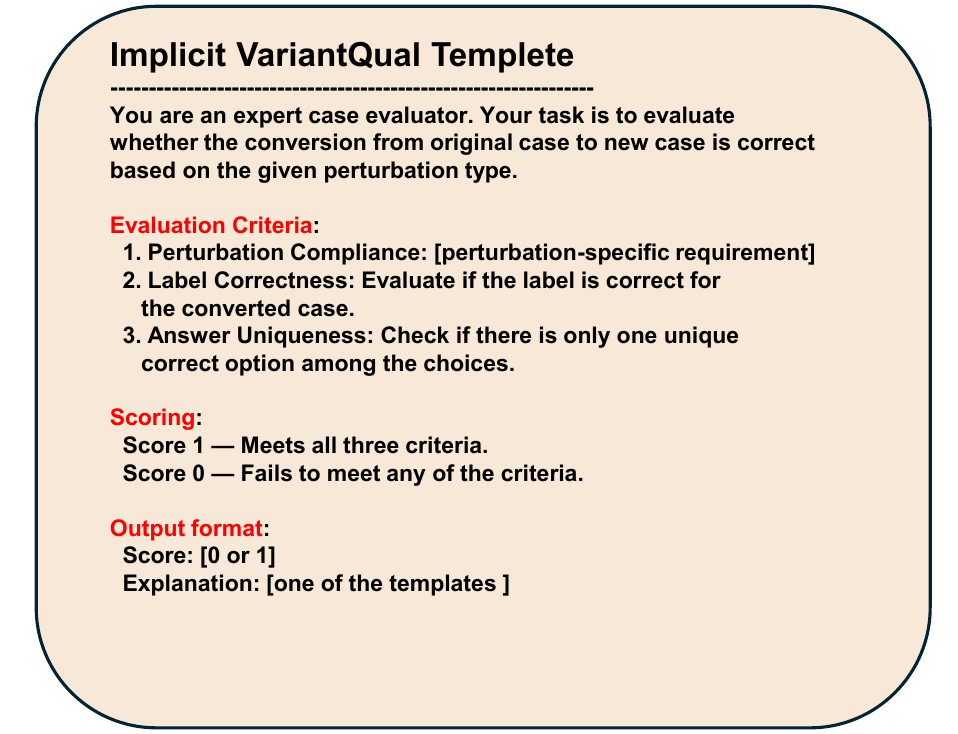}
  \caption{VariantQual implicit rubric prompt template.}
  \label{fig:prompt_implicitchecker}
\end{figure}

\begin{figure}[h]
  \centering
  \includegraphics[width=\linewidth]{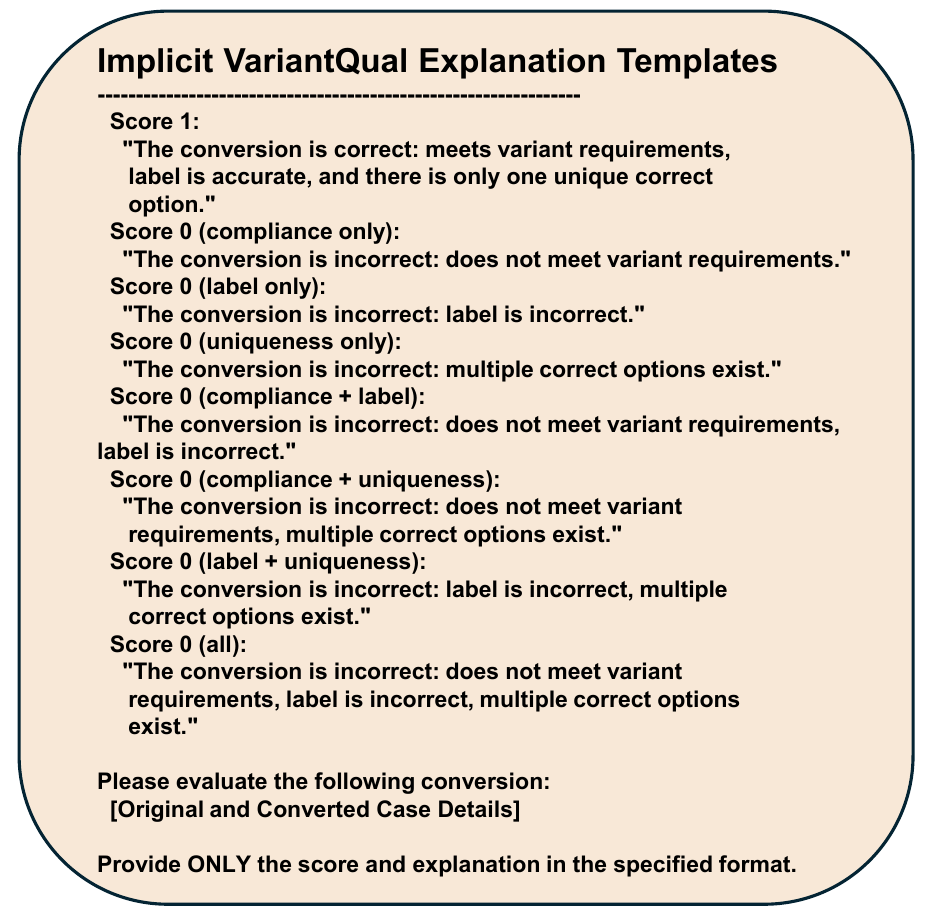}
  \caption{VariantQual implicit explanation prompt template.}
  \label{fig:implicitexplanation}
\end{figure}

\begin{figure}[h]
  \centering
  \includegraphics[width=\linewidth]{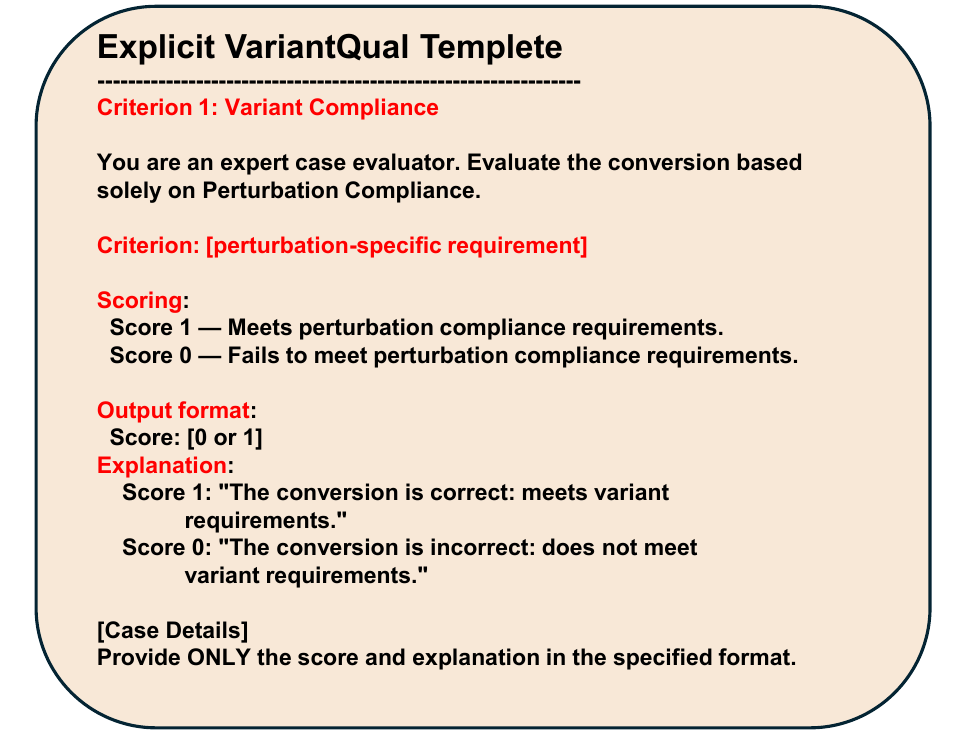}
  \caption{VariantQual explicit rubric prompt template.}
  \label{fig:prompt_explicitchecker}
\end{figure}

Figures~\ref{fig:prompt_generator_base}--\ref{fig:prompt_explicitchecker} show the complete prompt templates used in SAGE for the VariantGen and VariantQual components.

\section{Use of Large Language Models}
\label{app:llm_usage}

This work involves the use of LLMs in two capacities.
First, LLMs are integral to the research methodology itself: SAGE employs LLM-based components (VariantGen and VariantQual) for benchmark variant generation and quality verification, and uses proprietary LLMs such as Qwen2.5-Max as baselines for comparison experiments. These usages are documented in detail throughout Sections~\ref{sec:method} and~\ref{sec:experiments}.
Second, LLM-based writing assistants were used to polish and improve the clarity of the manuscript text.
The authors have carefully reviewed all content and take full responsibility for the correctness, originality, and integrity of the entire paper.



\end{document}